\documentclass{article}

\usepackage[final]{neurips_2023}

\usepackage[utf8]{inputenc} %
\usepackage[T1]{fontenc}    %
\usepackage{hyperref}       %
\usepackage{url}            %
\usepackage{booktabs}       %
\usepackage{amsfonts}       %
\usepackage{nicefrac}       %
\usepackage{microtype}      %
\usepackage{xcolor}         %

\usepackage{amsmath,amssymb}
\usepackage{mathtools}
\usepackage{amsthm}
\usepackage{listings}
\newcommand{\R}{\mathbf{R}}

\DeclareMathOperator{\vectorize}{\mathbf{vec}}
\theoremstyle{plain}

\theoremstyle{definition}

\theoremstyle{remark}

\usepackage{algorithmic, algorithm}

\title{A Negative Result on Gradient Matching\\ for Selective Backprop}

\author{%
Lukas Balles$^{1}$ \quad Cedric Archambeau$^{2*}$ \quad Giovanni Zappella$^1$ \\
$^1$Amazon Web Services \quad $^2$Helsing \\
\texttt{lukas.balles@gmail.com},
\texttt{cedric.archambeau@helsing.ai},
\texttt{zappella@amazon.de}
}

\begin{document}

\maketitle

\begin{abstract}
\let\thefootnote\relax\footnotetext{* Work done while at Amazon Web Services.}
With increasing scale in model and dataset size, the training of deep neural networks becomes a massive computational burden.
One approach to speed up the training process is \emph{Selective Backprop}.
For this approach, we perform a forward pass to obtain a loss value for each data point in a minibatch.
The backward pass is then restricted to a subset of that minibatch, prioritizing high-loss examples.
We build on this approach, but seek to improve the subset selection mechanism by choosing the (weighted) subset which best matches the mean gradient over the entire minibatch.
We use the gradients w.r.t.~the model's last layer as a cheap proxy, resulting in virtually no overhead in addition to the forward pass.
At the same time, for our experiments we add a simple random selection baseline which has been absent from prior work.
Surprisingly, we find that both the loss-based as well as the gradient-matching strategy fail to consistently outperform the random baseline.
\end{abstract}

\section{Introduction}
\label{sec:intro}

Deep learning models excel in a variety of applications, from natural language processing to computer vision.
With increasing scale in model and dataset size the training of such models becomes a massive computational burden.
This holds in particular for large \emph{foundation models} trained on massive amounts of data.
Thus, the research community is constantly striving to improve the efficiency of deep model training in various ways.

One notable approach is \emph{Selective Backprop} \citep{jiang2019accelerating}.
This technique performs a forward pass on a minibatch of data to obtain a loss value for each data point.
It then restricts the backward pass to a \emph{subset} of that minibatch, prioritizing points with high loss.
The intuition is that the gradient contributions of low-loss examples are small and therefore do not significantly influence the training process.
Since, according to commonly accepted estimates (e.g., see the estimates made by \cite{deepspeed-back}), the backward pass constitutes the roughly two thirds of the computational effort of a gradient computation, dropping a portion of the data after the forward pass can significantly speed up the training process.

In this work, we build on this approach and seek to improve the mechanism by which the subset of the minibatch is selected.
Instead of a simple prioritization of high-loss examples, we aim to select the subset which best preserves the mean gradient over the entire minibatch.
We use the gradients with respect to the model's last layer as a cheap and easy-to-compute proxy for the full gradients.
A sparse approximation algorithm may then be used to \emph{match} the mean gradient using a small, possibly weighted, subset of the minibatch.
This procedure can be implemented in a highly-efficient manner adding only minuscule overhead in addition to the forward pass.

This has the potential to result in more informative subsets.
For example, if a minibatch contains two near duplicates with high loss, a loss-based heuristic will likely select both instances for backpropagation.
Our matching approach will be aware of the redundancy and select only one of them.
It can also assign that instance a higher weight to account for the gradient contribution of its duplicate.

We evaluate our method on various image and text datasets using different architectures and optimization methods and find improvements over the loss-based sampling strategy of \citet{jiang2019accelerating}.
At the same time, we add a simple baseline that randomly selects a portion of the minibatch and was missing from the original work of \citet{jiang2019accelerating}.
Depending on the experimental setup, this is simply equivalent to training with a smaller batch size for the same number of steps or with the same batch size using fewer steps and an accelerated learning rate decay schedule.
In our experiments, Selective Backprop with either selection strategy fails to beat this simple baseline.
Our results confirm some of the findings in \citet{kaddour2023train} but we do not restrict our evaluation to model pre-training and we provide a quantification of the models performance with higher granularity.

\section{Selective Backprop}

Selective Backprop performs a forward pass of the model to obtain a loss value for each data point.
A point with loss $l\in \R$ is then included in the backward pass with a probability $p(l) = \text{CDF}(l)^\beta$, with the cumulative distribution function (CDF) of loss values across the \emph{entire} dataset as well as selectivity parameter $\beta>0$.
Larger values of $\beta$ imply a smaller selection probability.
Since the exact CDF would be too costly to recompute in each steps, \citet{jiang2019accelerating} approximate it by storing the $R$ most recent loss values seen during training.

The original paper proposed to keep performing forward passes and stochastically selecting points using $p(l)$ until $m$ points have been selected.
Then a backward pass is performed.
This leads to a variable number of forward-propagated points for each size-$m$-minibatch making it to the backward pass and is cumbersome to implement.
A slight variation of the method forward-propagates a fixed number of $M$ points, from which $m$ are selected.
In this case, the CDF can be approximated using the $M$ loss values and we set $\beta$ to be the inverse of the desired fraction, i.e., $\beta = M/m$.
This is the variant implemented in the popular \texttt{composer} library\footnote{\url{https://github.com/mosaicml/composer}} and we adopt this variant here.

An ideal implementation of Selective Backprop would cache the intermediate activations from the initial forward pass and reuse them in the backward pass.
As \citet{jiang2019accelerating} note, this is currently infeasible to implement in deep learning frameworks.
Therefore, they run a separate forward-backward pass on the selected subset, which we adopt for our experiments here.
\footnote{There may also be a (subtle) possible advantage to a separate forward pass.
The caching of intermediate activations is the most memory-intensive aspect of deep model training and, as a consequence, GPU memory limits the minibatch size that one can use.
Deactivating activation caching allows using a larger-than-usual minibatch size during the initial forward pass.
Combined with our a smart subset selection strategy, the resulting gradient estimate may become more accurate than what one could achieve with a randomly-sampled minibatch of a size dictated by GPU memory constraints.
}
\citet{jiang2019accelerating} note that the overhead of the initial \emph{selection} pass may be mitigated by reducing the floating point precision.

Since a backward pass is roughly twice as expensive as a forward pass \citep{jiang2019accelerating}, selective backprop with an extra selection pass incurs cost proportional to $\tfrac{M}{3} + m$, while backpropagating the full batch costs $M$.
Hence, we need to choose $m < \tfrac{2}{3}M$ to achieve cost savings.

\section{Related Work}
\label{sec:related_work}
While Selective Backprop ``prunes'' a given minibatch, another line of work seeks to sample more informative minibatches from the training set in the first place.
Multiple works \citep{alain2015variance,loshchilov2015online,katharopoulos2017biased,csiba2018importance} use importance sampling strategies to assemble minibatches.
These methods set the probability of selecting an example to a quantity proportional to an ``importance score''.
\citet{zhang2017determinantal} use determinantal point processes to sample \emph{diverse} minibatches.

Coreset methods select small subsets of large datasets with the goal that a model trained on the coreset retain, as closely as possible, the performance of a model trained on the entire dataset.
One of the criteria proposed by \citet{paul2021deep} is to select the data points having the highest gradient norm, either at initialization or after training for a few epochs.
The idea of gradient matching has been used for selecting coresets for deep learning models, among others, by \citet{killamsetty2021grad}. %
They also propose to use last-layer gradients as cheap approximations for full gradients, which we adopt in the present work.
Our proposed methods can be seen as applications of a gradient matching-based coreset methods to minibatches.
While coreset selection is done \emph{once}, we select subsets of each minibatch, demanding extreme efficiency.

\section{Method}

\label{sec:method}

The motivation for our method is simple.
Assume we are given a minibatch of size $M$ but only want to backpropagate $m<M$ points.
Ideally, we would like to choose the subset which best approximates the \emph{mean} gradient over the entire minibatch.
Of course, identifying that exact subset would require computing all individual gradients first, which would render the whole endeavor meaningless.
Instead, we use a cheap \emph{proxy} for these gradients, namely, the gradients with respect to the parameters of the \emph{last (linear) layer} of the model, which can be obtained at minuscule overhead in addition to the forward pass.

Denote the last-layer gradient of the $i$-th data point as $g_i$ and their mean as $\bar{g} = \tfrac{1}{M} \sum_{i=1}^M g_i$.
The \emph{weighted} subset of size $m< M$ which best preserves the mean gradient is given by
\begin{equation}
    \label{eq:cardinality_constrained_matching}
    \min_{w \in \R^M} \left\Vert \sum_{i=1}^M w_i g_i - \bar{g}\right\Vert^2 \quad \text{s.t.} \quad \Vert w \Vert_0 \leq m,
\end{equation}
where $\Vert\cdot\Vert_0$ denotes the $\ell_0$-pseudonorm which counts non-zero elements.
Eq.~\eqref{eq:cardinality_constrained_matching} is a cardinality-constrained quadratic optimization problem, which is NP-hard.
However, good approximate solutions can be obtained with greedy algorithms such as orthogonal matching pursuit \citep[OMP;][]{mallat1993matching}, which we briefly introduce in Appendix~\ref{app:matching_pursuit}.

We replace the loss-based sampling strategy in Selective Backprop by a weighted subset given by an approximate OMP solution to Eq.~\eqref{eq:cardinality_constrained_matching}.
Note that OMP can be implemented in a Gram-matrix variant, requiring only access to inner products $g_i^T g_j$.
This is \emph{crucial} to make our algorithm efficient, since the Gram matrix of the last-layer gradients can be computed efficiently with minuscule overhead compared to the forward pass, as we will explain in \textsection\ref{sec:last_layer_grads}.
The full algorithm is then described in \textsection\ref{sec:algorithm}.

\subsection{Obtaining Last-Layer Gradients}
\label{sec:last_layer_grads}

To facilitate our subset selection, we first need to extract the last-layer gradients.
As noted above, and discussed in detail in Appendix~\ref{app:matching_pursuit},
it suffices to compute the matrix $K\in\R^{M\times M}$ with $K_{ij} = g_i^T g_j$.
In the following, we explain how this Gram matrix of last-layer gradients can be computed \emph{implicitly} at virtually no overhead compared to the forward pass.

Assume the last layer has input dimension $D$ and output dimension $C$ and denote its inputs for each data point as $h_i \in \R^D$.
Let $W\in \R^{C\times D}$ be the weight matrix and $b\in \R^C$ the bias vector of the last layer, and let $L(\hat{y}, y)$ denote the loss function.
With that, we can define the loss for each data point as a function of the last layer's parameters:
\begin{equation}
    \label{eq:last_layer_loss}
    \ell_i(W, b) := L(Wh_i + b, y_i).
\end{equation}
Denote by $p_i = \nabla L(Wh_i + b, y_i) \in \R^C$ the gradient of the loss w.r.t.~the model output (its first argument).
With that, the gradients of the loss w.r.t.~$W$ and $b$ can be written as
\begin{equation}
    \nabla_W \ell_i(W, b) = p_i h_i^T, \quad \nabla_b \ell_i(W, b) = p_i,
\end{equation}
and the inner products are given by $K_{ij}  = \vectorize(p_i h_i^T)^T \vectorize(p_j h_j^T) + p_i^T p_j$, where $\vectorize(\cdot)$ denotes the vectorization of a matrix.
We show in Appendix~\ref{app:method_details} that this can be computed \emph{implicitly} as $K_{ij} = (h_i^T h_j) (p_i^T p_j) + p_i^T p_j$.
Hence, if we have access to the batch matrices $H = [h_1 \cdots h_M]^T \in \R^{M\times D}$ and
$P = [p_1 \cdots p_M]^T \in \R^{M\times C}$, we can compute $K$ as
\begin{equation}
    K = HH^T \circ PP^T + PP^T
\end{equation}
where $\circ$ denotes the elementwise product between matrices. Computing $K$ in this implicit fashion drastically reduces the computational cost and memory footprint compared to computing the individual last-layer gradients themselves.

In terms of a practical implementation in a deep learning framework, this requires (i) caching the last-layer inputs $H$ during the forward pass, and (ii) computing gradients $P$ of the loss w.r.t.~the model outputs.
The computational cost of the latter typically pales relative to the forward pass.
The following code snippet gives the gist of a PyTorch implementation.
\begin{lstlisting}
H = base_model(X)
output = output_layer(H)
loss = loss_fn(output, y)
P = torch.autograd.grad(loss, output)[0]
K = P @ P.T * (H @ H.T + 1.)
\end{lstlisting}
Full PyTorch code for our method may be found in the Appendix~\ref{app:code}.

\subsection{Full Algorithm}
\label{sec:algorithm}

As mentioned above, we follow the general strategy of \emph{Selective Backprop} while replacing the loss-based sampling with our gradient-matching selection.
Algorithm~\ref{alg:gradient_matching_minibatch} provides pseudocode.
We extract the Gram matrix of the last-layer gradients in an ``extended'' forward pass as described in the previous section.
We then use the Gram-based variant of OMP, which requires the Gram matrix $K$ as well as the vector $t\in \R^M$ with $t_i = g_i^T\bar{g}$, see Appendix~\ref{app:matching_pursuit}.
The latter can easily be computed by taking row-wise mean over $K$.
OMP returns a set of active indices $I\subset [M]$ with $\vert I \vert = m$ as well as a vector of associated weights $\gamma \in \R^m$.
These weights might not sum to $1$, resulting in a scaled gradient.
We have found it to be beneficial to normalize these weights.\footnote{
We adopt the convention that non-weighted methods shall correspond to a weight vector containing ones, summing to $m$.
Therefore, we multiply the weights by $m$ in Alg.~\ref{alg:gradient_matching_minibatch}.}

In Eq.~\eqref{eq:cardinality_constrained_matching}, we do not enforce any constraints on the weights.
In principle, it would be possible for OMP to assign negative weights, even though that seems unlikely and did not occur in our experiments.
As a safeguard, one can simply clip the weights at zero.
Alternatively, one could employ algorithms for non-negative matching pursuit, which we leave for future work.
In a similar vein, one might consider regularizing the weights to avoid over-reliance on individual data points, which we did not pursue further at this time.

\begin{algorithm}\small
  \caption{Gradient Matching Minibatch}
  \label{alg:gradient_matching_minibatch}
\begin{algorithmic}
  \INPUT{Minibatch $X\in\R^{M\times D_\text{in}}$, $y\in \R^{M\times D_\text{out}}$, desired subset size $m$},
  \STATE $K = \texttt{extended\_forwad\_pass}(X, y)$
  \STATE $t = \frac{1}{M} \sum_{i=1}^M K_{:,i} \in \R^M$
  \STATE $I, \gamma = \texttt{OMP}(K, t, m)$
  \STATE Normalize $\gamma \gets m \gamma / \Vert \gamma\Vert_1$.
  \OUTPUT{$I$, $\gamma$}
\end{algorithmic}
\end{algorithm}

\section{Experiments}
\label{sec:experiments}

We now present an empirical evaluation of our method on various deep learning tasks.
We compare our gradient matching approach to the original loss-based sampling strategy of \citet{jiang2019accelerating}.
We also add a simple baseline of randomly selecting subsets without replacement, which could be see as an equivalent for having a smaller batch size.

\paragraph{Models and Datasets}
We run experiments on a total of five dataset/model combinations.
We train a ResNet-18 \citep{he2016deep} on both CIFAR-10 and CIFAR-100 \citep{krizhevsky2009learning}, a WideResNet-16-4 \citep{zagoruyko2016wide} on the SVHN dataset \citep{netzer2011reading}, a WideResNet-28-2 \citep{zagoruyko2016wide} on a $32\times 32$px variant of ImageNet \citep{chrabaszcz2017downsampled}, as well as fine-tuning a pretrained Bert model \citep{devlin2018bert} on the IMDB dataset~\citep{imdb-data}.

\paragraph{Training Settings}
For each model and dataset, we chose default settings with a base batch size as well as optimizer and learning rate schedule.
Details may be found in Appendix~\ref{app:experiment_details}.
The training budget and learning rate schedules are defined in number of epochs, which we count in terms of the \emph{forward} passes, as done by \citet{jiang2019accelerating}.
This results in a reduction of the total number of backpropagated data points by the subsampling fraction, which we denote as $\rho$ here.
For the batch size, we experimented with two settings.
\emph{Fixed batch size:} We always use the same base batch $M \equiv M_\text{base}$ size during the forward pass, irrespective of the subsampling fraction.
Consequently, for all $\rho$, we use the same number of update steps but different effective batch sizes $m = \rho M_\text{base}$.
\emph{Scaled batch size:} We scale the batch size used during the forward pass as $M = M_\text{base} / \rho$, such that $m\equiv M_\text{base}$ irrespective of $\rho$.
Consequently, all $\rho$ use the same effective batch size but the number of update steps is reduced by a factor of $\rho$.
The latter strategy is in line with the work of \citet{jiang2019accelerating}, who always collect points until $M_\text{base}$ is reached.
Since the results did not differ substantially, we only report results for the fixed batch size, which we believe is more in line with how such a method would be used in practice.
Results for the scaled batch size can be found in Appendix~\ref{app:experiment_details}.

\subsection{Main Results}

\begin{figure}
    \centering
    \includegraphics[width=0.32\textwidth]{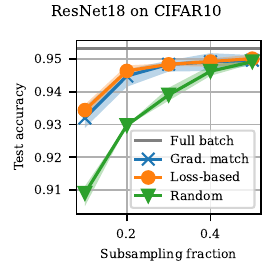}
    \includegraphics[width=0.32\textwidth]{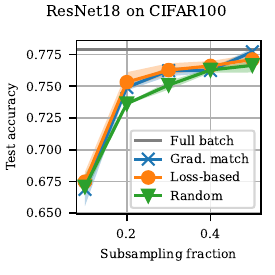}
    \includegraphics[width=0.32\textwidth]{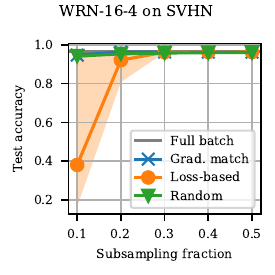}\\
    \includegraphics[width=0.32\textwidth]{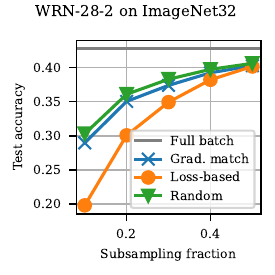}
    \includegraphics[width=0.32\textwidth]{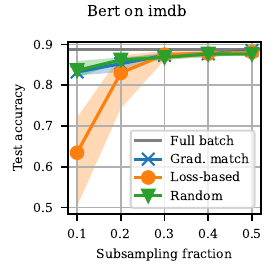}
    \caption{Maximal test accuracy achieved by each subsampling strategies at varying subsampling fractions.
    Results are averaged over three random seeds and the shaded area spans minimal/maximal values observed.}
    \label{fig:main_results}
\end{figure}

Figure~\ref{fig:main_results} depicts our main results.
In addition to the loss-based strategy of \citet{jiang2019accelerating}, we add our gradient-matching approach as well as the random baseline.
The plots show the test accuracy achieved by each method for different values of the subsampling fraction $\rho$.
We see that the gradient matching approach consistently matches or outperforms the loss-based sampling strategy, especially for lower subsampling fractions.
However, either method fails to consistently outperform the simple random baseline.

In the experiments above, all methods use the same learning rate schedule and are assigned the same budget.
Since the learning rate has been tuned to the base batch size, one might worry that this skews the results.
In Appendix~\ref{app:experiment_details}, we tune the learning rate separately for each method and find similar results.

\subsection{Results With Label Noise}

\begin{figure}
    \centering
    \includegraphics[width=0.32\textwidth]{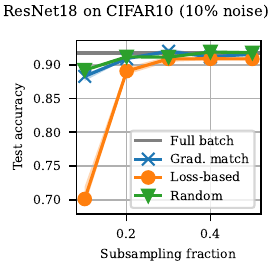}
    \includegraphics[width=0.32\textwidth]{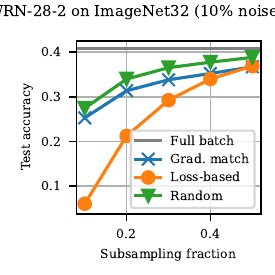}
    \includegraphics[width=0.32\textwidth]{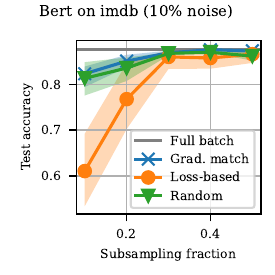}
    \label{fig:noisy_results}
    \caption{Results with $10\%$ label noise.
    Maximal test accuracy achieved by each subsampling strategies at varying subsampling fractions.
    Results are averaged over three random seeds and the shaded area spans minimal/maximal values observed.}
\end{figure}

\citet{jiang2019accelerating} note as a potential shortcoming of their method that it might be sensitive to outliers or mislabeled examples.
Since such data points tend to have high loss, the loss-based selection will oversample them.
We repeated our experiments while adding label noise, i.e., we reassign random labels to a portion of the training data.
Figure~\ref{fig:noisy_results} shows results with $10\%$ label noise on three datasets, additional results may be found in Appendix~\ref{app:experiment_details}.
We see that the loss-based strategy suffers significantly compared to the case without label noise.

\subsection{Quality of the Gradient Estimate}

\begin{figure}
    \centering
    \includegraphics[width=0.32\textwidth]{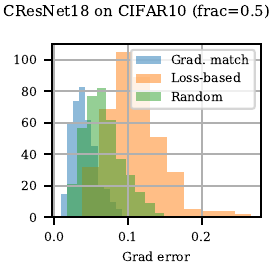}
    \includegraphics[width=0.32\textwidth]{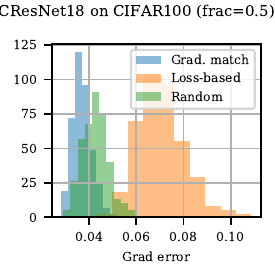}
    \includegraphics[width=0.32\textwidth]{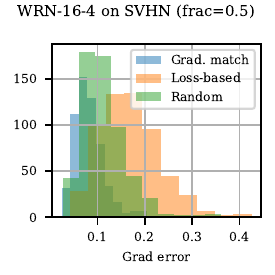}
    \caption{Histogram of squared errors of the gradient estimate for different subsampling strategies.}
    \label{fig:grad_errors}
\end{figure}

One goal of selective backprop should be to achieve better gradient estimates than a randomly sampled minibatch of the same size.
Here we compare the quality of the gradient estimate for the different subsampling strategies.
Given model weights, we compute the true gradient over the entire datasets.
Then we sample a number of minibatches, subsample each with the respective strategy, and compute the squared $L_2$ distance to the full-dataset gradient.
Figure~\ref{fig:grad_errors} shows histograms for three datasets with gradients computed at a random initialization of the model.
We see that the gradient matching technique \emph{does} achieve lower gradient errors than random subsampling while the loss-based sampling strategy even inflates the error.

\section{Conclusion}

We have investigated a gradient-matching approach to improve the selection mechanism in the \emph{Selective Backprop} framework of \citet{jiang2019accelerating}.
This technique can outperform the existing loss-based sampling strategy, especially in the presence of label noise.
Moreover, we have shown that the gradient approximation obtained using this method is better than the one obtained using a random sample of data.
Nevertheless, our experiments show that Selective Backprop fails to ouperform random selection.

\bibliography{references}
\bibliographystyle{icml2022}

\newpage
\appendix

\section{Matching Pursuit}
\label{app:matching_pursuit}

Orthogonal matching pursuit \citep{mallat1993matching} approximately solves problems of the form
\begin{equation}
    \label{eq:omp_problem_general}
    \min_{w\in\R^{M}} \Vert Aw - b\Vert^2 \quad\text{s.t.}\quad \Vert w \Vert_0 \leq m,
\end{equation}
That is, it matches a target signal $b\in \R^D$ by a sparse weighted sum $Aw = \sum_i w_i a_i$ of so-called atoms $a_i\in \R^D$.
Solving this problem exactly is NP-hard.

We give a brief description of OMP here, for details refer to the original paper.
We represent a subset using a list of indices $I\subset [M]$ and associated weights $\gamma \in \R^{\vert I \vert}$.
A given $I$ and $\gamma$ yields an approximation $A_I\gamma$, where $A_I$ denotes the restriction of $A$ to columns $I$.
Matching pursuit greedily adds the element which best matches the residual, i.e., it chooses $k_\ast = \text{arg\,max}_{k \in [M]} a_k^T (b - A_I\gamma)$.
After adding a new element to $I$, orthogonal matching pursuits recomputes the weights $\gamma$ such as to minimize $\Vert A_I\gamma - b\Vert^2$, which results in $\gamma = (A_I^TA_I)^{-1} A_I^T b$.

Both the selection criterion as well as the formula for $\gamma$ depend solely on inner products $a_i^Ta_j$ as well as $a_i^Tb$.
Therefore, OMP can be performed given only access to the matrix $K = A^TA$ as well as the vector $t=A^Tb$.

Algorithm~\ref{alg:omp_gram} provides pseudo-code.
In practice, the inversion of $K_{I, I}$ need not be done from scratch in each iteration.
Rather, the algorithm maintains a Cholesky factorization of $K_{I,I}$, which can be updated efficiently upon the addition of a new element.
We refer to the work of \citet{rubinstein2008efficient} for details.
For our experiments, we used the implementation in \texttt{scikit-learn} \citep{scikit-learn}.

\begin{algorithm}[H]
  \caption{Orthogonal Matching Pursuit (Gram variant)}
  \label{alg:omp_gram}
\begin{algorithmic}
  \INPUT{$K\in\R^{M\times M}$, $t\in\R^{M}$}, desired subset size $m$
  \STATE Initialize $\alpha = t$
  \WHILE{$\vert I \vert < m$}
  \STATE Select $k=\text{arg\,max}_{i\in [M]} \alpha_i$
  \STATE $I\gets I \cup \{k\}$
  \STATE $\gamma \gets K_{I,I}^{-1}t_I$
  \STATE $\alpha \gets \alpha - K_{:,I}\gamma$
  \ENDWHILE
  \OUTPUT{$I, \gamma$}
\end{algorithmic}
\end{algorithm}

\section{Method Details}
\label{app:method_details}

\subsection{Mathematical Details}
In \textsection\ref{sec:last_layer_grads}, we have used the identity
\begin{equation}
    \vectorize(p_1 h_1^T)^T \vectorize(p_2 h_2^T) = (h_1^T h_2) (p_1^T p_2)
\end{equation}
for $p_1, p_2\in\R^C$ and $h_1, h_2\in\R^D$.

\begin{proof}
We denote elements of the vectors above using double indices, e.g., $p_{1i} = [p_1]_i$.
\begin{equation}
    \begin{split}
        \vectorize(p_1h_1^T)^T \vectorize(p_2h_2^T) & = \sum_{i, j} [p_1h_1^T]_{ij} [p_2h_2^T]_{ij} = \sum_{i, j} (p_{1i}h_{1j})(p_{2i}h_{2j}) \\
        & = \left( \sum_i p_{1i}p_{2i} \right)\left( \sum_j h_{1j}h_{2j} \right) = (p_1^Tp_2)(h_1^T h_2).
    \end{split}
\end{equation}
\end{proof}

\subsection{PyTorch Code}
\label{app:code}

In the following, we show PyTorch code for our implementation of gradient matching for selective backprop.

\begin{lstlisting}
import numpy as np
import torch


class GradMatchSelector(torch.nn.Module):

    def __init__(self, model, output_layer, loss_fn, k):
        self._model = model
        self._output_layer = output_layer
        self._loss_fn = loss_fn
        self._k = k
        
        # Cache inputs to output layer during forward pass.
        self._last_layer_input = None
        def cache_hook(module, inputs, output):
            self._last_layer_input = inputs[0]
        self._hook = output_layer.register_forward_hook(cache_hook)

    def _compute_last_layer_grad_gram_matrix(self, X, y):
        output = self._model(X)
        loss = self._loss_fn(output, y).sum()
        output_grads = torch.autograd.grad(loss, output)[0]
        PPt = output_grads @ output_grads.T
        HHt = self._last_layer_input @ self._last_layer_input.T
        return PPt * (HHt + 1.)
    
    def forward(self, X, y):
        Gram = self._compute_last_layer_grad_gram_matrix(X, y)
        # Use existing numpy-based implementation of OMP.
        Gram = Gram.cpu().numpy().astype(np.float64)
        Xy = Gram.sum(1)
        weights, idx, _ = _gram_omp(Gram, Xy, self._k)
        # Convert weights to torch and normalize.
        weights = torch.from_numpy(weights).float().to(X.device)
        weights = torch.clamp(weights, min=0.)
        weights /= weights.sum()
        weights *= len(idx)
        return idx, weights
\end{lstlisting}

\subsection{Computational Cost}

Table~\ref{tab:epoch_times} shows the wall-clock time per training epoch for our method compared to the purely loss-based Selective Backprop.
The current implementation of our approaches adds a small overhead (less than 5\%).
A more detailed profiling of the run time indicates that most of this overhead stems from the GPU $\to$ CPU transfer of the Gram matrix $K$, which could be avoided with a PyTorch implementation of (Gram) orthogonal matching pursuit.
We leave that for future work.

\begin{table}[H]
    \caption{Wall-clock time [s] per epoch for vanilla selective backprop (SB) and gradient matching (GM).}
    \centering
    \footnotesize
    \begin{tabular}{llrrr}
\toprule
         & method &  GM &    SB \\
dataset & fraction &    &       \\
\midrule
CIFAR10 & 0.1 & 20.70 & 19.69 \\
         & 0.3 & 27.41 & 26.77 \\
         & 0.5 & 36.74 & 35.68 \\
CIFAR100 & 0.1 & 20.37 & 19.95 \\
         & 0.3 & 27.05 & 26.56 \\
         & 0.5 & 36.28 & 35.21 \\
\bottomrule
\end{tabular}

    \label{tab:epoch_times}
\end{table}

\section{Experimental Details and Additional Results}
\label{app:experiment_details}

\subsection{Hyperparameter Settings}
The experiments used the following hyperparameter settings:
\begin{itemize}
    \item CIFAR-10(0): We use SGD with Nesterov momentum of 0.9, weight decay of $5\cdot 10^{-4}$. We train for 200 epochs, starting with a learning rate of $0.1$ and decaying by a factor of $0.2$ after $60$, $120$ and $160$ epochs. The base batch size is $128$.
    \item SVHN: We use SGD with Nesterov momentum of 0.9, weight decay of $5\cdot 10^{-4}$. We train for $80$ epochs with a Cosine Annealing schedule. The initial learning rate is $0.01$ and the base batch size is $128$.
    \item ImageNet-32: We use SGD with momentum of $0.9$, weight decay of $5\cdot 10^{-4}$. We train for $40$ epochs, starting with a learnign rate of $0.01$ and decaying by a factor of $0.2$ after $10$, $20$ and $30$ epochs. The base batch size is $128$.
    \item IMDB: We use AdamW with default values for $\beta_1$ and $\beta_2$ and weight decay of $5\cdot 10^{-4}$.
    We fine-tune the pretrained Bert model for $3$ epochs using a constant learning rate of $2\cdot 10^{-5}$.
    The base batch size is $32$.
\end{itemize}

\subsection{Results with Scaled Base Batch Size}

\begin{figure}[H]
    \centering
    \includegraphics[width=0.32\textwidth]{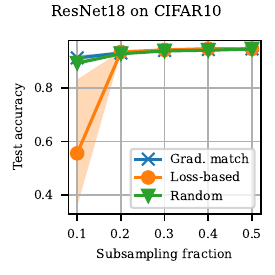}
    \includegraphics[width=0.32\textwidth]{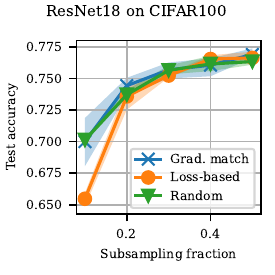}
    \includegraphics[width=0.32\textwidth]{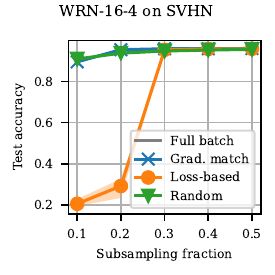}\\
    \includegraphics[width=0.32\textwidth]{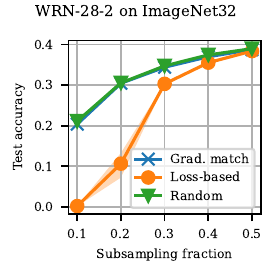}
    \includegraphics[width=0.32\textwidth]{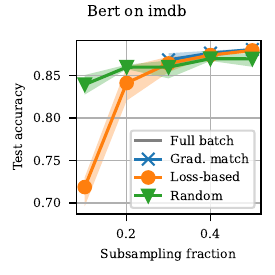}
    \caption{Results under the base learning rate schedule with a scaled base batch size. We show the maximal test accuracy achieved by each method at varying subsampling fractions.}
    \label{fig:scaledbs_results}
\end{figure}

Figure~\ref{fig:scaledbs_results} shows results for the scaled base batch size, as explained in Section~\ref{sec:experiments}.
That is, the forward pass uses a batch size of $M=M_\text{base} / \rho$.
The results do not differ substantially from the alternative batch size setting presented in the main text.
All methods achieve lower accuracy on average, suggesting that more steps with a smaller batch size are preferable to fewer steps with a large batch size.

\subsection{Additional Results with Label Noise}

\begin{figure}[H]
    \centering
    \includegraphics[width=0.32\textwidth]{fig/gradmatch-notune__CIFAR10__0__ResNet18__maxacc.pdf}
    \includegraphics[width=0.32\textwidth]{fig/gradmatch-notune__CIFAR10__10__ResNet18__maxacc.pdf}
    \includegraphics[width=0.32\textwidth]{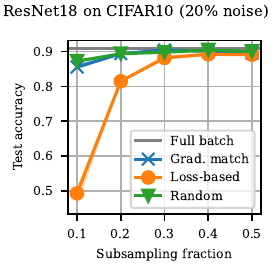}\\
    \includegraphics[width=0.32\textwidth]{fig/gradmatch-notune__CIFAR100__0__ResNet18__maxacc.pdf}
    \includegraphics[width=0.32\textwidth]{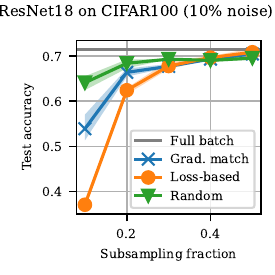}
    \includegraphics[width=0.32\textwidth]{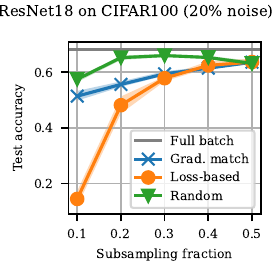}\\
    \includegraphics[width=0.32\textwidth]{fig/gradmatch-notune__SVHN__0__WRN-16-4__maxacc.pdf}
    \includegraphics[width=0.32\textwidth]{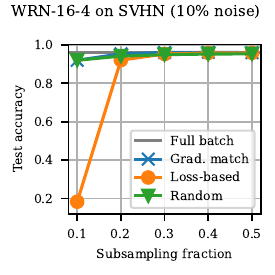}
    \includegraphics[width=0.32\textwidth]{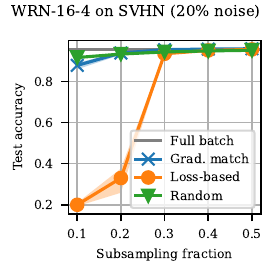}\\
    \includegraphics[width=0.32\textwidth]{fig/gradmatch-notune__ImageNet32__0__WRN-28-2__maxacc.pdf}
    \includegraphics[width=0.32\textwidth]{fig/gradmatch-notune__ImageNet32__10__WRN-28-2__maxacc.pdf}
    \includegraphics[width=0.32\textwidth]{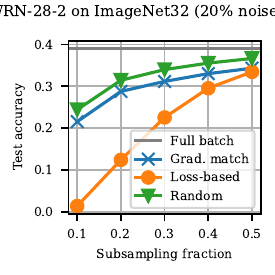}\\
    \includegraphics[width=0.32\textwidth]{fig/gradmatch-notune__imdb__0__Bert__maxacc.pdf}
    \includegraphics[width=0.32\textwidth]{fig/gradmatch-notune__imdb__10__Bert__maxacc.pdf}
    \includegraphics[width=0.32\textwidth]{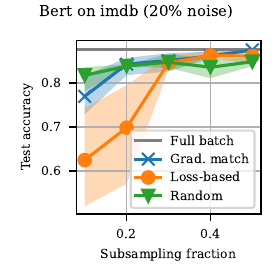}
    \label{fig:all_noisy_results}
\end{figure}

Figure~\ref{fig:all_noisy_results} shows results with various levels of label noise.

\subsection{Learning Rate Tuning}

In addition to the experiments with the base learning rate schedule, we conducted experiments where learning rate schedules are tuned separate for each method.
The rationale is that the subsampling techniques affect the effective batch size used to compute gradients and it is well known the batch size and step size are highly interdependent hyperparameters. %
Therefore, using a learning rate schedule tuned to the base batch size may skew the results.

To keep the experimental workload manageable, we use a simplified tuning protocol that is based on the base schedule and a ``linear scaling rule'' for the relationship between batch size and learning rate.
Under this rule, originally proposed in the distributed deep learning community, a doubling of the batch size (for randomly sampled batches) would necessitate a doubling of the learning rate as well.
Hence, for each method and subsampling fraction, we tune a learning rate factor $\beta \in [0, 1]$ and set the (initial) learning rate to $\alpha = \beta \cdot \alpha^\text{base}$.
We then use this factor to ``stretch'' the learning rate schedule by setting the total number of epochs to $T = T^\text{base} / \beta$.

One can think of $\beta$ as the ``effective'' downsampling fraction of a method.
If the downsampling were done uniformly at random, we would expect a value of $\beta = m/M$ to be optimal, as per the linear scaling rule.
For smarter subset selections strategies, we hope to achieve values of $\beta$ that are significantly larger than that.
We tune $\beta$ on a logarithmic grid spanning the interval $10^{-1}$ to $10$.
A value of $1$ recovers the base case.

\begin{figure}[H]
    \centering
    \includegraphics[width=0.32\textwidth]{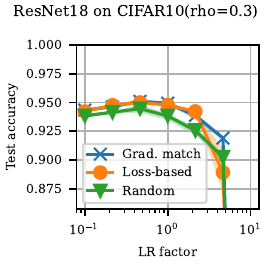}
    \includegraphics[width=0.32\textwidth]{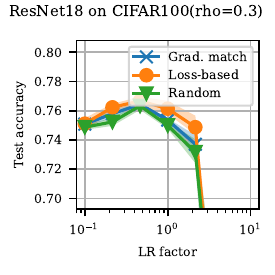}
    \includegraphics[width=0.32\textwidth]{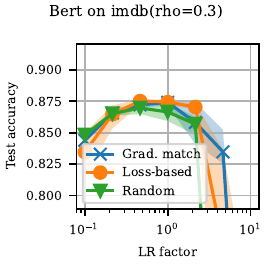}
    \caption{Sensitivity when tuning the learning rate. We show the maximal test accuracy achieved by each method as a function of $\beta$, which stretches the learning rate as well as the budget.}
    \label{fig:tuned_results}
\end{figure}

Results are depicted in Fig.~\ref{fig:tuned_results} in the form of a sensitivity plot.
We show results for $\rho=0.3$ here; additional values may be found in Appendix~\ref{app:experiment_details}.

\end{document}